\documentclass[11pt]{article}
\usepackage{acl2016}
\usepackage{times}
\usepackage{url}
\usepackage{latexsym}
\usepackage{subcaption}
\usepackage{graphicx}
\usepackage{breqn}
\usepackage{float}
\usepackage{hyperref}
\usepackage{amsfonts}
\usepackage{comment}
\usepackage{multirow}
\usepackage{array}
\usepackage{tabularx}
\usepackage{csvsimple}

\usepackage[font=small,labelfont=bf]{caption}

\aclfinalcopy 

\title{Investigating how well contextual features are captured by bi-directional recurrent neural network models}

\author{Kushal Chawla$^{1}$$^*$, Sunil Kumar Sahu$^{2}$\thanks{$^*$Part of this work was done while authors were students at IIT Guwahati.}, Ashish Anand$^{3}$ \\
$^{1}$Adobe Research, Big Data Experience Lab, Bangalore, Karnataka, India \\
$^{2}$National Center for Text Mining, The University of Manchester, United Kingdom\\
$^{3}$Department of Computer Science and Engineering, IIT Guwahati, Assam, India\\
{\tt kchawla@adobe.com}\\
{\tt sunil.sahu@manchester.ac.uk}\\
{\tt anand.ashish@iitg.ernet.in}\\
}

\date{}

\begin{document}
\maketitle
\begin{abstract}
Learning algorithms for natural language processing (NLP) tasks traditionally rely on manually defined relevant contextual features. On the other hand, neural network models using an only distributional representation of words have been successfully applied for several NLP tasks. Such models learn features automatically and avoid explicit feature engineering. Across several domains, neural models become a natural choice specifically when limited characteristics of data are known. However, this flexibility comes at the cost of interpretability. In this paper, we define three different methods to investigate ability of bi-directional recurrent neural networks (RNNs) in capturing contextual features. In particular, we analyze RNNs for sequence tagging tasks. We perform a comprehensive analysis on general as well as biomedical domain
datasets. Our experiments focus on important contextual words as features, which can easily be extended to analyze various other feature types. We also investigate positional effects of context words and show how the developed methods can be used for error analysis.
\end{abstract}

\section{Introduction} 
Learning approaches for NLP tasks can be broadly put into two categories based on the way features are obtained or defined. 
The traditional way is to design features according to a specific problem setting and then use appropriate learning approach. Examples of such methods include classification algorithms like SVM~\cite{hong2005relation} and CRF~\cite{lafferty2001conditional} among others for several NLP tasks. A significant proportion of overall effort is spent on feature engineering itself. The desire to obtain better performance on a particular problem makes the researchers come up with a domain and task-specific set of features. The primary advantage of using these models is their interpretability. However, dependence on handcrafted features limits their applicability in low resource domain where obtaining a rich set of features is difficult.

On the other hand, neural network models provide a more generalised way of approaching problems in NLP domain. The models can learn relevant features with minimal efforts in explicit feature engineering. This ability allows the use of such models for problems in low resource domain.

The primary drawback of neural network models is that they are too complicated to interpret as the features are not manually defined.
Neural networks have been applied significantly to various tasks without many insights on what the underlying structural properties are and how the models learn to classify the inputs correctly. Mostly inspired by computer vision~\cite{simonyan2013deep,nguyen2015deep}, several mathematical and visual techniques have been developed in this direction~\cite{elman1989representation,karpathy2015visualizing,li2016understanding}.

In contrast to the existing works, this study aims to investigate ability of recurrent neural models to capture important context words. Towards this goal, we define multiple measures based on word erasure technique~\cite{li2016understanding}. We do a comprehensive analysis of performance of bi-directional recurrent neural network models for sequence tagging tasks using these measures. Analysis is focused at understanding how well the relevant contextual words are being captured by different neural models in different settings. The analysis provides a general tool to compare between different models, show that how neural networks follow our intuition by giving importance to more relevant words, study positional effects of context words and provide error analysis for improving the results.

\section{Proposed Methods}
A sequence tagging task involves assigning a tag (from a predefined set) to each element present in a given sequence. We model Name Entity Recognition (NER) as a sequence tagging task. We follow BIO-tagging scheme, where each named entity type is associated with two labels, $B-entity$ (standing for \textit{Beginning}) and $I-entity$ (standing for \textit{Intermediate}). The BIO scheme uses another label $O$(standing for {\it Other}) for all the context or non-entity words.

In this section, we discuss three methods to calculate the importance score of context words. Each method creates a different ranking of context words corresponding to each entity type for a given dataset. The methods range from simple frequency based to considering sentence level or individual word level effects. We assume that we have a pretrained model $M$ on a given dataset.

\subsection{Based on word frequency}
For a given sentence $S$ $\in$ test set $D$, consider a window of a particular size around each entity phrase (single or multi word, defined by true tags) $w_{e}$ in $S$. We increment the score (corresponding to  $w_{e}$'s entity type $e$ only) for each of the context words present in this window by one. For instance, the CoNLL-2003 shared task data (described in section~\ref{sec:dataset}) has 4 entity types, namely, {\it organization ($ORG$)}, {\it location ($LOC$)}, {\it person ($PER$)} and {\it miscellaneous ($MISC$)}. The corresponding labels under BIO-tagging scheme are {\it B-ORG}, {\it I-ORG}, {\it B-LOC}, {\it I-LOC} and so on. For a 2-word phrase with true tags as ({\it B-LOC}, {\it I-LOC}), the score corresponding to $LOC$ for each context word (with true tag as $O$) in the window is incremented by one. Let the score for a context word $w_{c}$ corresponding to entity type $e$ in one sentence be $A(w_{c},e,S)$.

Hence the relevance score is calculated as follows:

\begin{equation}
I(w_{c},e) = \frac{\sum\limits_{\forall S \in D} A(w_{c},e,S)}{\sum\limits_{\forall w_{c}}\sum\limits_{\forall S \in D} A(w_{c},e,S)}
\end{equation}

Using inverse frequency to account for irrelevant, too frequent words, the score can be calculated as follows:

\small
\begin{dmath}
I(w_{c},e) = \left(\frac{\sum\limits_{\forall S \in D} A(w_{c},e,S)}{\sum\limits_{\forall w_{c}}\sum\limits_{\forall S \in D} A(w_{c},e,S)}\right) \left(\frac{\sum\limits_{\forall e^'}\sum\limits_{\forall w_{c}}\sum\limits_{\forall S \in D} A(w_{c},e^',S)}{\sum\limits_{\forall e^'}\sum\limits_{\forall S \in D} A(w_{c},e^',S) + k}\right)
\end{dmath}
\normalsize

where $k$ accounts for 0 counts and sum over $e^'$ means summing over all the remaining entity types. In our experiments, we use $k$=1 and a window size of 11 (5 words on each side). We refer to these methods collectively as M\_WF in rest of the paper.

\subsection{Using sentence level log likelihood}
In the M\_WF method, the relevance of each context word is calculated irrespective of its dependence on other words in the sentence. We define another measure using sentence level log likelihood to take into account the dependency between words in a sentence. We refer to this method as M\_SLL in rest of the paper.

Let the set of all context words be $W$ and that of all entity types be $E$. Define $S_{w_{c},e}$ as the set of all sentences where both the word $w_{c}$ $\in$ $W$ and entity type $e$ $\in$ $E$ are present. We say that an entity type $e$ is present in a sentence $S$, if $\exists$ a word $\in$ $S$ which has it's true tag corresponding to entity type $e$. Let $F(w_{c},e)$ be the size of set $S_{w_{c},e}$.

Now, let the true tag sequence for a sentence $S$ be $S_{TAGS}$. For a context word $w_{c}$ $\in$ $S$,  let $L_{1}(w_{c},S)$ be the negative log likelihood of $S_{TAGS}$ obtained from pretrained model $M$. Note that since we are working at a sentence level, $L_{1}(w_{c},S)$ will be same for all the context words and entities present in $S$.

We adapt the erasure method of ~\newcite{li2016understanding}. Here, we replace the representation of word $w_{c}$ with a random word representation having same number of dimensions and recalculate the negative log likelihood for the true tag sequence $S_{TAGS}$. Let this value be $L_{2}(w_{c},S)$. Intuitively, if $S \in S_{w_{c},e}$ and $w_{c}$ is relevant for the entity type $e$, the probability of the true sequence should decrease when the word is removed from the sentence. Correspondingly, it's negative log likelihood value should increase. Hence, the score $I(w_{c},e)$ for a given word corresponding to the entity type can be calculated in the following manner:

\small
\begin{equation}
I(w_{c},e) = \frac{1}{F(w_{c},e)} \sum\limits_{\forall S \in S_{w_{c},e}} \frac{L_{2}(w_{c},S) - L_{1}(w_{c},S)}{L_{1}(w_{c},S)}
\end{equation}
\normalsize

\subsection{Considering left and right word contexts separately}
The relevance scoring method M\_SLL does not distinguish between words present in the same sentence. The third method, referred to as M\_LRC, works at word level and calculates relevance score of each word by distinguishing its presence in the left or right side of the entity word. The measure is defined in a way that it does take into account of dependency between words in the sentence. In a bi-directional setting, the hidden layer representation for any word in a sentence, is a concatenation of two representations - one which combines words to the left, and the other which combines the words to the right. 

In the output layer, we combine the weight parameters and the hidden layer representation by a dot product. We divide this dot product in two parts as discussed below. 
Say the hidden representation is $h$ and weight parameters corresponding to a tag t $\in$ $T$ (set of all possible tags) are represented by $p_{t}$. We can write the dot product $p^{T}_{t}h$ as a sum of two dot products $p^{T}_{t,L}h_{L}$ and $p^{T}_{t,R}h_{R}$, representing the contribution from left and right parts separately. In our experiments, we also include the bias term as a weight parameter.

Now, take a sentence $S$, a context word $w_{c}$ in $S$, and an entity word $w_{e}$ in $S$ with true tag $t$ $\in$ $T$ corresponding to entity type $e$ $\in$ $E$. 
Define $AvgSum(w_{c},w_{e},S)$ as follows:

\small
\begin{equation}
AvgSum(w_{c},w_{e},S) = \frac{\sum\limits_{\forall f \in T-\{t\}} p^{T}_{f,K}h_{K}}{\alpha}
\end{equation}
\normalsize

where $\alpha$ is the size of the set $T - \{t\}$ and $K$ is either $L$ or $R$ depending on whether the word $w_{c}$ lies to the left or right of $w_{e}$ respectively. Notice that this sum is over all the false tags in set $T$ for the word $w_{e}$.

With the intuition that the important word should have higher dot product corresponding to true tag than to false tags, we define the score $L_{1}(w_{c},w_{e},S)$ as follows:

\small
\begin{dmath}
L_{1}(w_{c},w_{e},S) = \frac{ p^{T}_{t,K}.h_{K} - AvgSum(w_{c},w_{e},S)}{AvgSum(w_{c},w_{e},S)}
\end{dmath}
\normalsize

We again employ word erasure technique and recompute the above score by replacing the representation of word $w_{c}$ with a random word representation. We call it $L_{2}(w_{c},w_{e},S)$. Now, we can compute the final score for this instance $L(w_{c},w_{e},S)$ as:

\small
\begin{dmath}
L(w_{c},w_{e},S) = \frac{L_{1}(w_{c},w_{e},S) - L_{2}(w_{c},w_{e},S)}{L_{2}(w_{c},w_{e},S)}
\end{dmath}
\normalsize

The relevance score $I(w_{c},e)$ is then computed by taking average of $L(w_{c},w_{e},S)$ over all instances.

\section{Experiments}
We consider the task of sequence tagging problem for evaluation and analysis of the proposed methods to interpret neural network models. In particular, we choose the three variants of recurrent neural network models for Named Entity Recognition(NER) task.

\subsection{Model architecture}
\begin{figure}[h]
\centering
\includegraphics[width=0.5\textwidth]{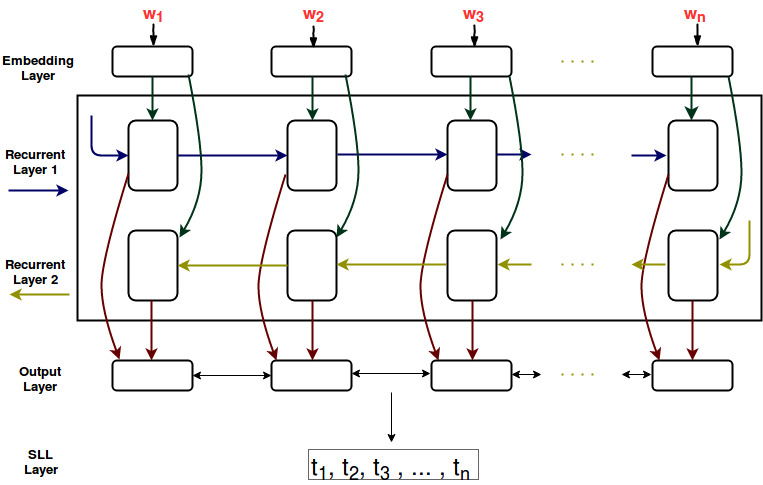}
\caption{General model architecture for a bi-directional recurrent neural network in sequence tagging problem.}
\label{fig:model}
\end{figure}

The generic RNN model architecture used for this work is given in figure \ref{fig:model}.

Input layer contains all the words in the sentence. In the embedding layer, each word is represented by it's $d$ dimensional vector representation. The hidden layer contains a bi-directional recurrent neural network which outputs a $2h$ dimensional representation for every word, where $h$ is the number of hidden layer units in the recurrent neural network. In bi-directional models, both the past and future contexts are used to represent the words in a given sentence. Finally, a fully connected network connects the hidden layer to the output layer, which contains scores for each possible tag corresponding to every word in the sentence. A sentence level log likelihood loss function~\cite{collobert2011natural} is used in the training process.

 For this work, we experiment with standard bi-directional Recurrent Neural Network (Bi-RNN), bi-directional Long Short Term Memory Network (Bi-LSTM)~\cite{graves2013generating,huang2015bidirectional} and bi-directional Gated Recurrent Unit Network(Bi-GRU)~\cite{chung2014empirical}. For simplicity, we refer to these bi-directional models as RNN, LSTM and GRU in rest of the paper. 
 
\begin{table*}[t]
\begin{minipage}{\textwidth}
\centering
\scalebox{0.9}{
\begin{tabular}  
{|p{0.15\linewidth}|p{0.1\linewidth}|p{0.1\linewidth}|p{0.1\linewidth}|p{0.1\linewidth}|p{0.1\linewidth}|p{0.1\linewidth}|p{0.1\linewidth}|} 
\hline
\multirow{2}{*}{\bf Dataset} & \multicolumn{3}{c|}{\bf Instances} & \multicolumn{4}{c|}{\bf Test Set Performance} \\ \cline{2-8}
 &{\bf Training} & {\bf Validation} & {\bf Testing} & {\bf Model} & {\bf Precision} & {\bf Recall} & {\bf F Score} \\ \hline
\multirow{3}{*}{CoNLL-2003} & \multirow{3}{*}{14987} & \multirow{3}{*}{3466} & \multirow{3}{*}{3684} & RNN & 83.42 & 81.77 & 82.59 \\  \cline{5-8}
& & & & LSTM & 85.87 & 84.41 & {\bf 85.13} \\ \cline{5-8} 
& & & & GRU & 85.11 & 83.66 & 84.38 \\ \cline{1-8} 
\hline
\multirow{3}{*}{JNLPBA-2004} & \multirow{3}{*}{18046} & \multirow{3}{*}{500} & \multirow{3}{*}{3856} & RNN & 67.71 & 68.99 & 68.34 \\  \cline{5-8}
& & & & LSTM & 67.94 & 72.69 & {\bf 70.23} \\ \cline{5-8} 
& & & & GRU & 67.55 & 70.05 & 68.78 \\ \cline{1-8} 
\end{tabular}
}
\caption{Statistics and performance of different models on two NER datasets used in this work.}
\label{tab:baseline}
\end{minipage}
\end{table*}

\subsection{Datasets}
\label{sec:dataset}
In this work, we use two NER datasets from diverse domains. One is from generic domain whereas other is from biomedical domain. Statistics of both datasets are given in Table~\ref{tab:baseline}.

\textbf{CoNLL, 2003}: This dataset was released as a part of CoNLL-2003 language independent named entity recognition task~\cite{tjongkimsang2003conll}. Four named entity types have been used: location, person, organization and miscellaneous. For this work, we have used the original split of the English dataset. There were 8 tags used {\it I-PER}, {\it B-LOC}, {\it I-LOC}, {\it B-ORG}, {\it I-ORG}, {\it B-MISC}, {\it I-MISC} and $O$. We focus on three entity types, namely, location ($LOC$), person ($PER$) and organization ($ORG$) in our analysis. For this dataset, we use pretrained GloVe 50 dimensional word vectors~\cite{pennington2014glove}.

\textbf{JNLPBA, 2004}: Released as a part of Bio-Entity recognition task~\cite{kim2004introduction} at JNLPBA in 2004, this dataset is from GENIA version 3.02 corpus~\cite{kim2003genia}. There are 5 classes in total - {\it DNA}, {\it RNA}, {\it Cell\_line}, {\it Cell\_type} and {\it Protein}. We use all the classes in our analysis. There are 11 tags, 2 (for begin and intermediate word) for each class and $O$ for other context words. We use 50 dimensional word vectors trained using skip-gram method on a biomedical corpus~\cite{mikolov2013efficient,mikolov2013distributed}.
For this work, we calculate the relevance scores for all the words which have their true tag as $O$ for any test instance in the two datasets.

\subsection{Correlation measures}
In the output (last) layer we take dot product between weight parameters and the hidden layer outputs and expect that this value (normalized) would be highest corresponding to the true tag. To obtain these similarities between distributions of hidden layer outputs to the weight parameters, we consider two other measures apart from dot product:
\begin{enumerate}
\item \textbf{Kullback-Leibler Divergence}: Given two discrete probability distributions \textbf{A} and \textbf{B}, the Kullback-Leibler Divergence(or KL Divergence) from \textbf{B} to \textbf{A} is computed in the following manner:
\begin{equation}
D_{KL}(A||B) = \sum\limits_{i} A(i) \log \frac{A(i)}{B(i)}
\label{eq:KL}
\end{equation}
$D_{KL}(A||B)$ may be interpreted as a measure to see that how good the distribution \textbf{B} approximates the distribution \textbf{A}. For our experiments, we take normalized weight parameters as \textbf{A} and hidden representations as \textbf{B}. The lower this KL-divergence is, higher is the correlation between \textbf{A} and \textbf{B}.

\item \textbf{Pearson Correlation Coefficient}: Given two variables \textbf{X} and \textbf{Y}, Pearson Correlation Coefficient(PCC) is defined as:
\begin{equation}
\rho_{X,Y} = \frac{cov(\textbf{X},\textbf{Y})}{\sigma_{X}\sigma_{Y}}
\end{equation}
where $cov(\textbf{X},\textbf{Y})$ is the covariance, $\sigma_{X}$ and $\sigma_{Y}$ are the standard deviations of \textbf{X} and \textbf{Y} respectively. $\rho_{X,Y}$ takes the values between -1 and 1. 
\end{enumerate}

\section{Results and Discussion}

Throughout our experiments, we use \textbf{50} dimensional word vectors, \textbf{50} hidden layer units, learning rate as \textbf{0.05}, number of epochs as \textbf{21} and a batch size of \textbf{1}. The performance of various models on both the datasets is summarized in Table \ref{tab:baseline}. Among the three bi-directional models, LSTM performs the best.

\subsection{Correlation Analysis}
\begin{figure}[h]
\centering
\includegraphics[width=0.5\textwidth]{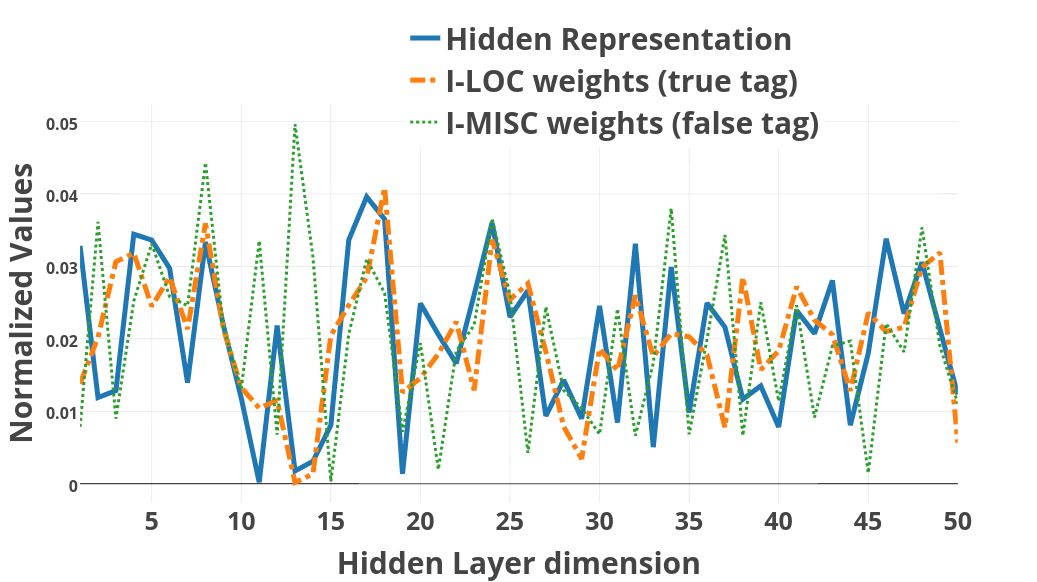}
\caption{Visualization of hidden representation of a $LOC$ entity word ``Rangoon'' and weight parameters corresponding to true and false tags.}
\label{fig:visualize}
\end{figure}

We analyze the correlation between the hidden layer representations and the weight parameters connecting hidden and output layers. Meeting our expectation, this correlation of hidden layer values is found to be higher with the weight parameters corresponding to the true tag for a given input word. For instance, take a sentence from ConLL dataset: ``The students, who had staged an 11-hour protest at the junction in northern Rangoon, were taken away in three vehicles.''. Here, the word ``Rangoon'' has it's true tag as {\it I-LOC} and rest all are context words. Figure \ref{fig:visualize} plots the normalized values for left side part of the hidden representation for ``Rangoon'', along with corresponding weight parameters for {\it I-LOC} and {\it I-MISC} tags.
 {\it I-MISC} has been chosen as it's corresponding dot product is maximum among all the false tags. The high correlation between the hidden representation and weight parameters for the true tag can be clearly observed from the figure.
 
Table \ref{tab:Corr} gives the correlation values for above three measures corresponding to the ``Rangoon'' instance.
\begin{table}[h]
\centering
\scalebox{0.75} {
\begin{tabular} {|c|c|c|c|}
\hline
\textbf{Tag} & {\bf Dot Product} & {\bf KL Divergence} & {\bf PCC}  \\ \hline
I-LOC (True tag) & \textbf{7.27} & \textbf{0.15} & \textbf{0.62}\\ \hline
I-MISC (False Tag) & 1.76 & 0.48 & 0.17\\ \hline
\end{tabular}
}
\caption{Correlation values obtained corresponding to ``Rangoon'' instance from CoNLL dataset.}
\label{tab:Corr}
\end{table}

\subsection{Analysis of Relevance Scores}
In order to evaluate the ability of RNN models to capture important contextual words, we do a qualitative analysis at both word and sentence levels. This section provides instances from both CoNLL and JNLPBA datasets to illustrate how the three measures can be used to identify salient words with respect to bi-directional model. Although we compute word rankings using the three measures described above, our demonstrations in the paper primarily focus on the M\_LRC method. M\_LRC is able to treat each word individually with due attention to dependency on another words in a given sentence.

At the word level, we further breakdown the visualizations into three types:

\begin{figure*}[h]
\begin{subfigure}{.33\textwidth}
  \centering
  \includegraphics[width=\linewidth]{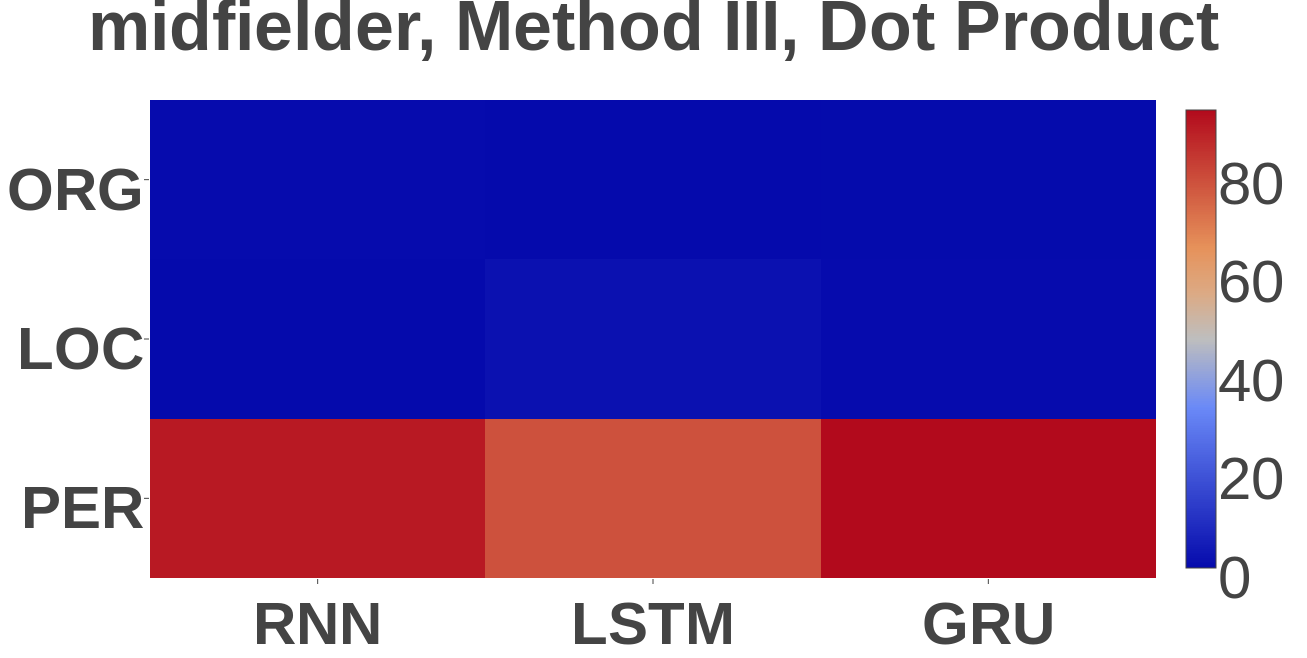}
  \caption{}
  \label{fig:md}
\end{subfigure}
\begin{subfigure}{.33\textwidth}
  \centering
  \includegraphics[width=\linewidth]{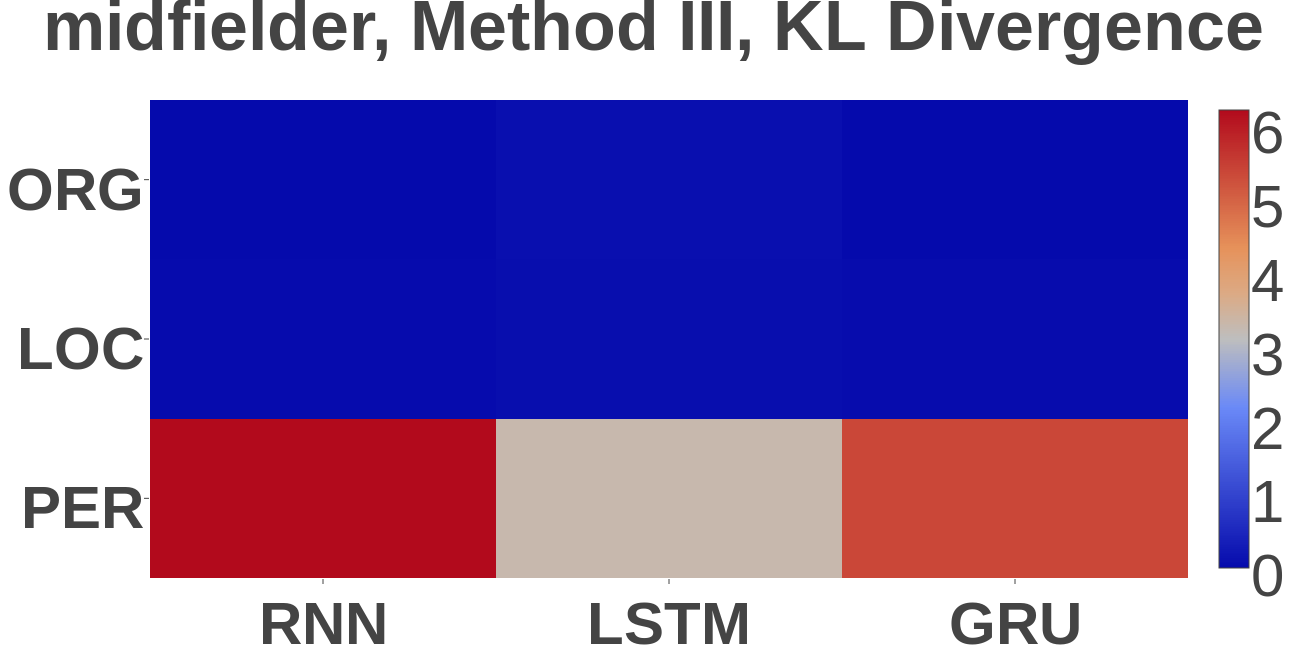}
  \caption{}
  \label{fig:mk}
\end{subfigure}
\begin{subfigure}{0.33\textwidth}
  \centering
  \includegraphics[width=\linewidth]{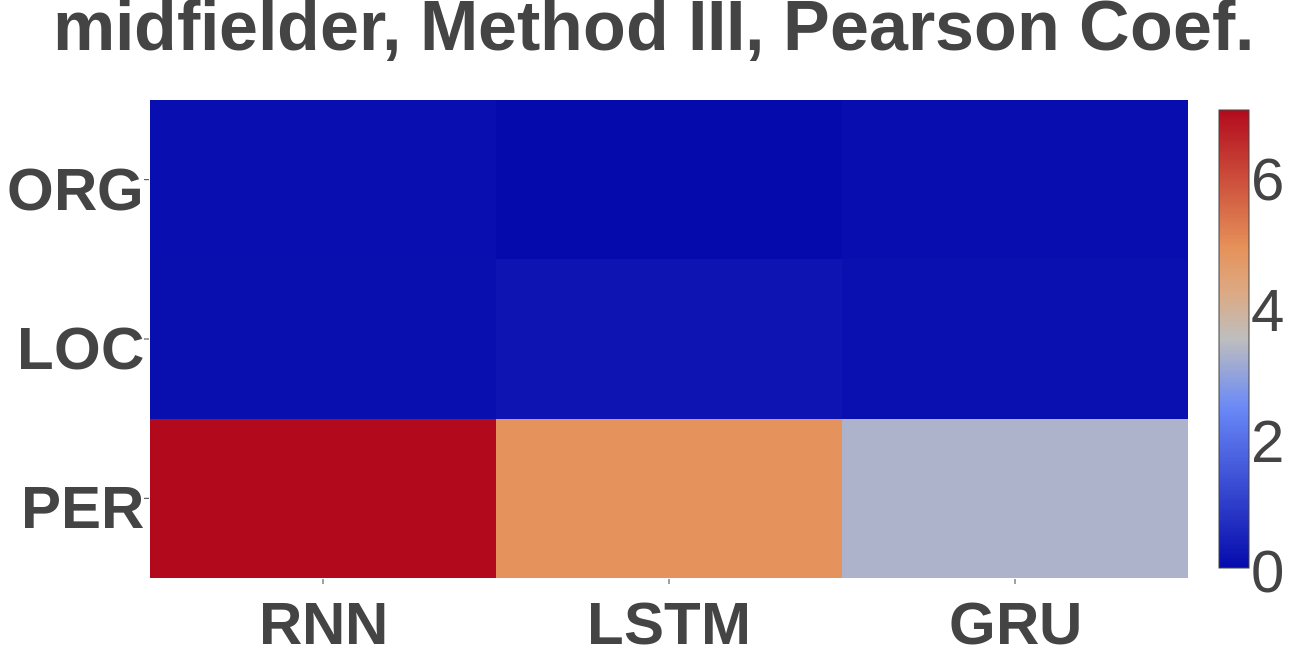}
  \caption{}
  \label{fig:ml}
\end{subfigure}

\begin{subfigure}{.33\textwidth}
  \centering
  \includegraphics[width=\linewidth]{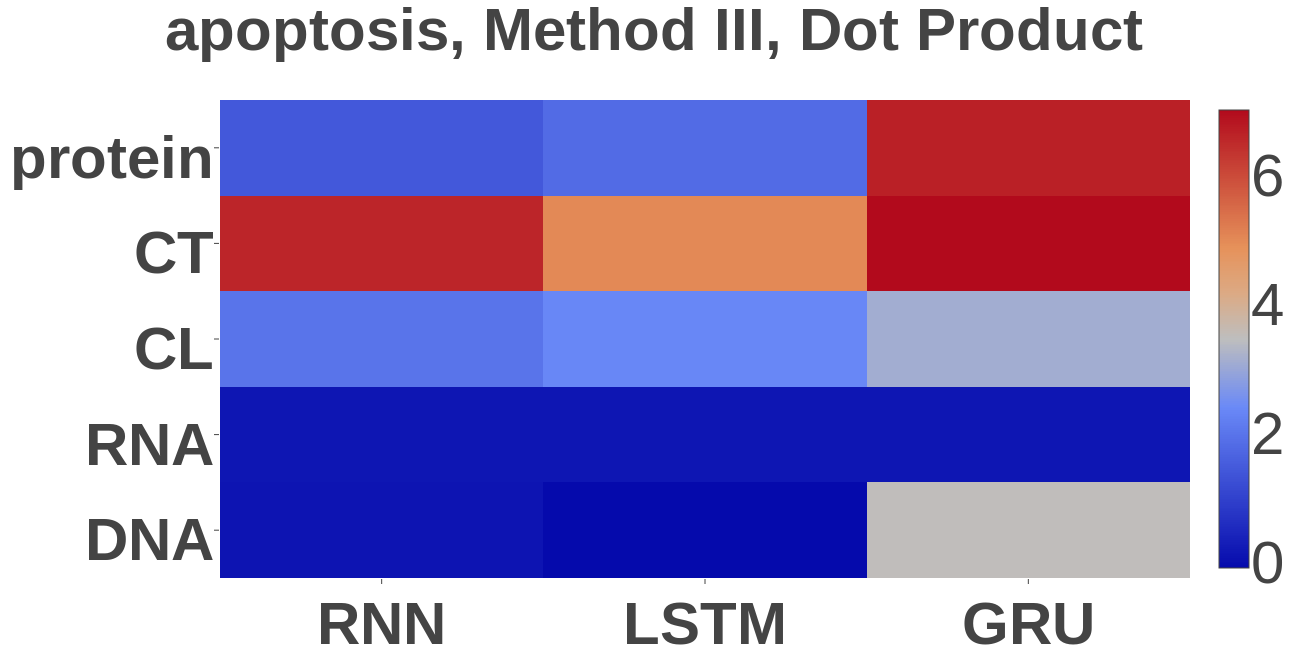}
  \caption{}
  \label{fig:j1}
\end{subfigure}
\begin{subfigure}{.33\textwidth}
  \centering
  \includegraphics[width=\linewidth]{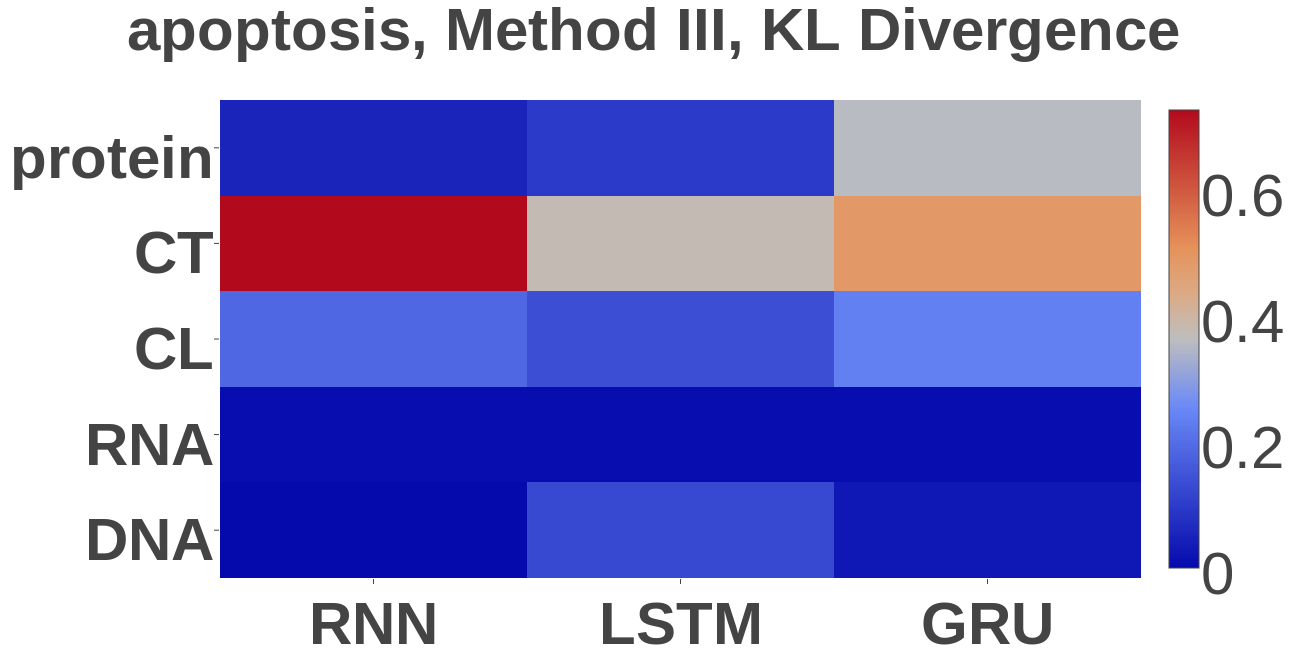}
  \caption{}
  \label{fig:j2}
\end{subfigure}
\begin{subfigure}{0.33\textwidth}
  \centering
  \includegraphics[width=\linewidth]{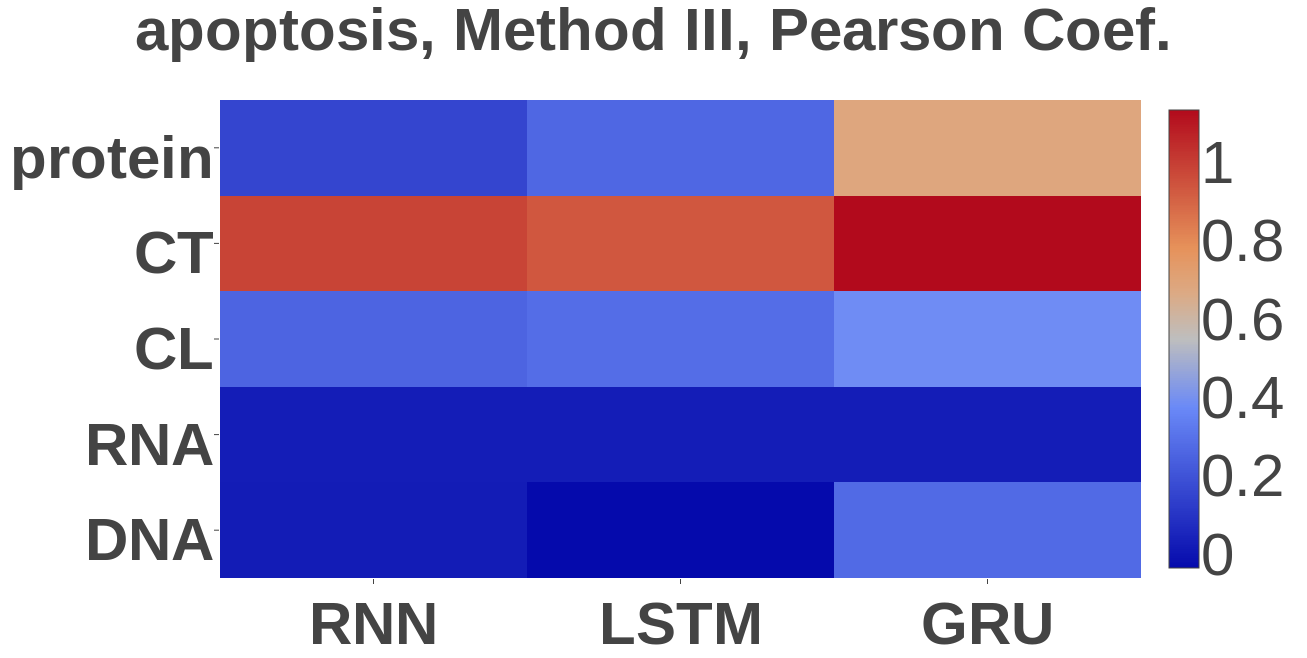}
  \caption{}
  \label{fig:j3}
\end{subfigure}

\caption{Heatmaps showing the scores for different words across models, entities and methods on CoNLL dataset in part (a), (b) and (c) and on JNLPBA dataset in (d), (e) and (f). Here, CT refers to $cell\_type$ and CL refers to $cell\_line$.}
\label{fig:midfielder_heatmaps}
\end{figure*}


\textbf{Fixing a word and a method}: In this case, we fix a particular word and use M\_LRC method. We analyze how the importance scores change with various models, entities and correlation measures. Figures \ref{fig:md}, \ref{fig:mk} and \ref{fig:ml} show heatmaps by fixing the word ``midfielder'' and M\_LRC method for CoNLL dataset. Based on our intuition, the word ``midfielder'' should have higher importance scores for $PER$ entity. This is clearly visible in the illustrations. All the three correlation measures are able to capture this intuition to a reasonable extent. Similarly, figures \ref{fig:j1}, \ref{fig:j2} and \ref{fig:j3} show heatmaps for ``apoptosis'' on JNLPBA dataset. The higher scores given to class $CT$ ({\it cell\_type}) are in agreement with the results of M\_WF method as well as with our intuition as ``apoptosis" indicates cell death. It can also be observed that all the bidirectional models do quite well in both these cases.

\begin{figure*}[h]
\begin{subfigure}{.33\textwidth}
  \centering
  \includegraphics[width=\linewidth]{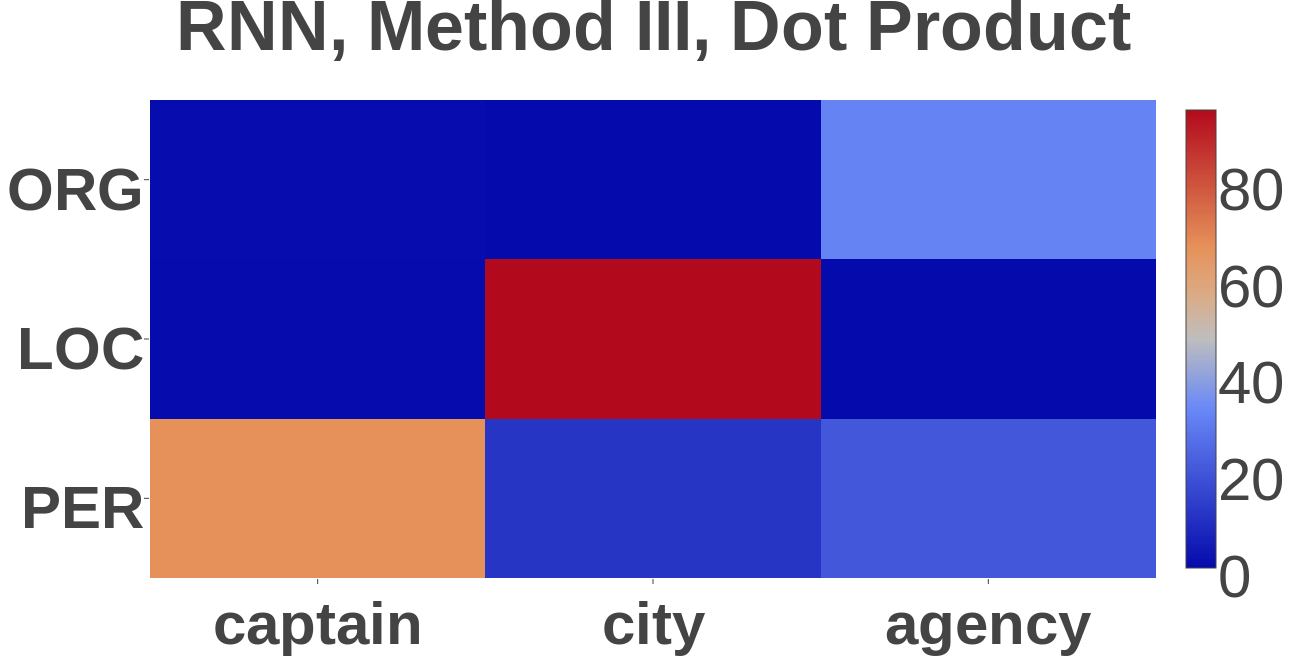}
  \caption{}
  \label{fig:dr}
\end{subfigure}
\begin{subfigure}{.33\textwidth}
  \centering
  \includegraphics[width=\linewidth]{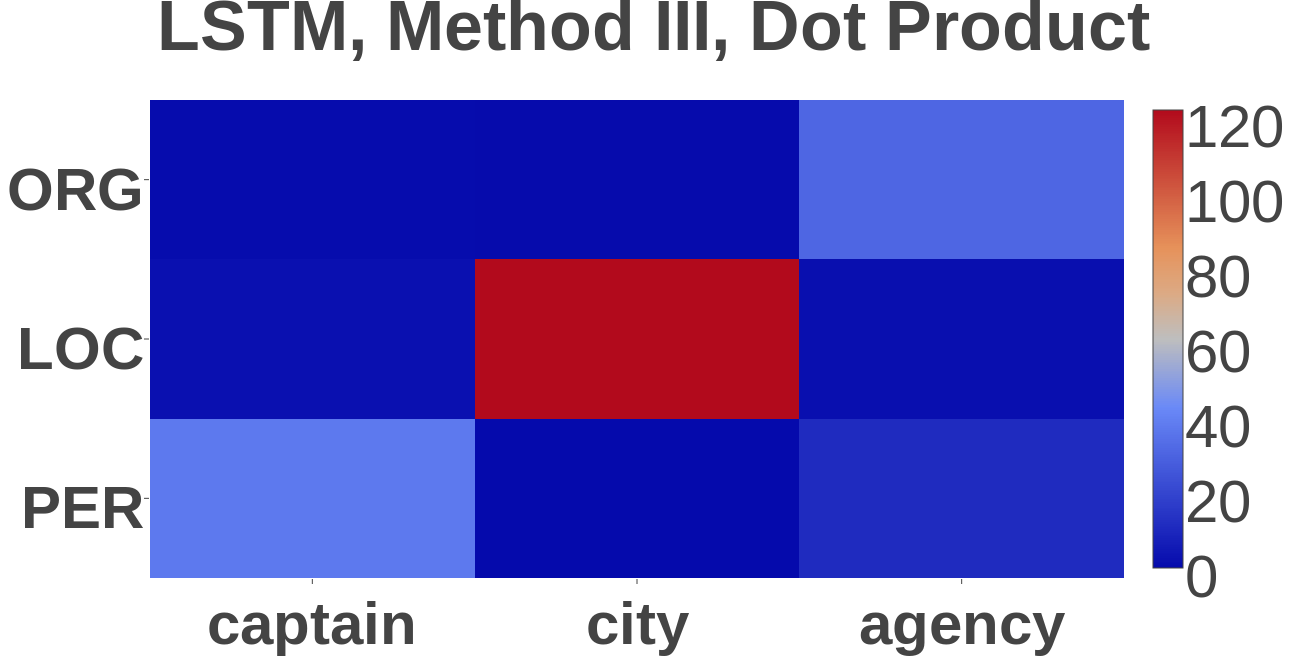}
  \caption{}
  \label{fig:dl}
\end{subfigure}
\begin{subfigure}{0.33\textwidth}
  \centering
  \includegraphics[width=\linewidth]{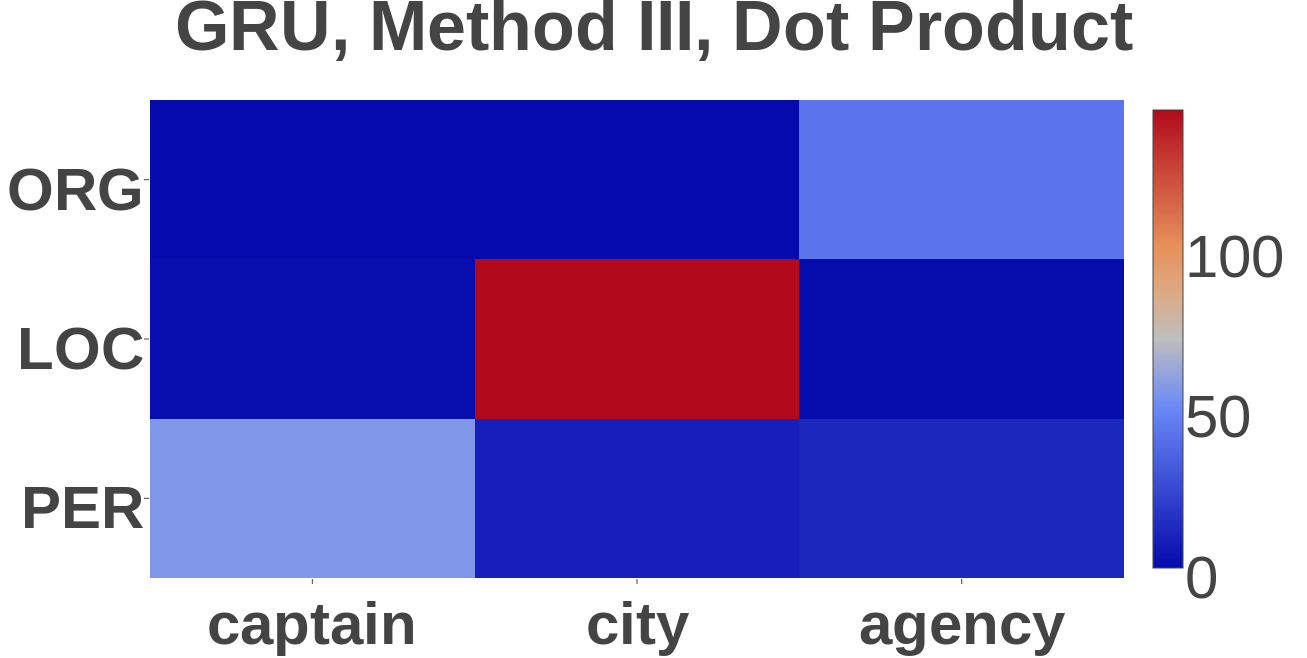}
  \caption{}
  \label{fig:dg}
\end{subfigure}

\caption{Heatmaps showing the word scores fixing a model with M\_LRC method using dot product on CoNLL dataset.}
\label{fig:dot_models}
\end{figure*}


\textbf{Fixing a model and a method}: In this case, we fix a particular model and try to visualize how the models score different contextual words for different entity types. Figure \ref{fig:dot_models} shows the heatmaps by fixing RNN, LSTM and GRU respectively with M\_LRC method (using dot product). Our intuition that ``captain'', ``city'' and ``agency'' would be relevant for $PER$, $LOC$ and $ORG$ entities respectively, is proved to be true as can be observed in all of the cases. However, neural models are unable to associate ``agency'' with $ORG$ as distinctively as in case of ``captain'' and ``city''. This can be attributed to frequent occurrence of the word ``agency'' in the context of words belonging to $PER$ or $LOC$ entities, thereby, confusing the models.

\begin{figure*}[h]
\begin{subfigure}{.33\textwidth}
  \centering
  \includegraphics[width=\linewidth]{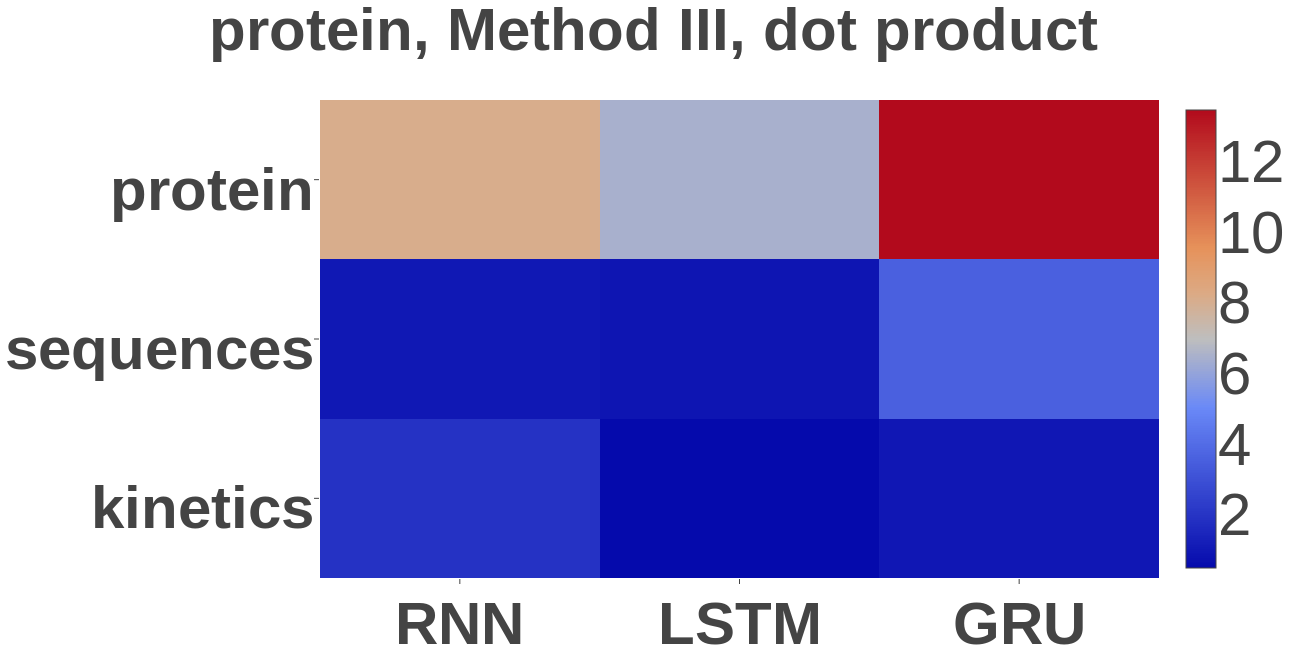}
  \caption{}
  \label{fig:dpp}
\end{subfigure}
\begin{subfigure}{.33\textwidth}
  \centering
  \includegraphics[width=\linewidth]{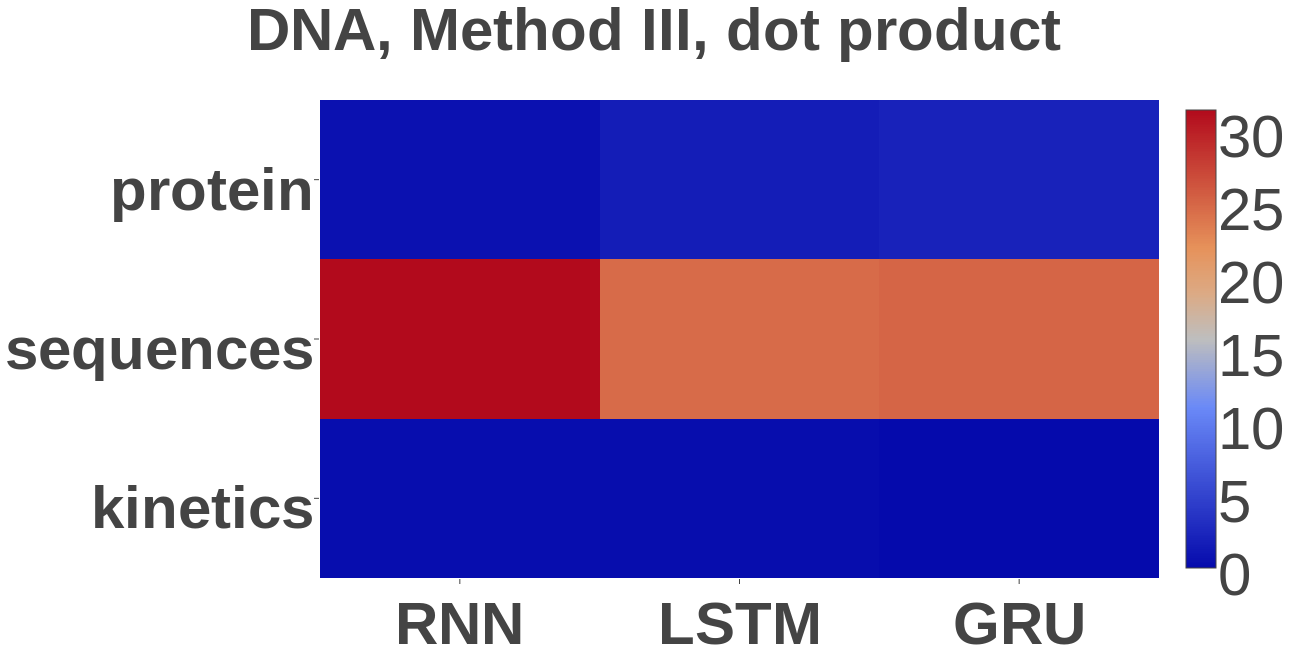}
  \caption{}
  \label{fig:dll}
\end{subfigure}
\begin{subfigure}{0.33\textwidth}
  \centering
  \includegraphics[width=\linewidth]{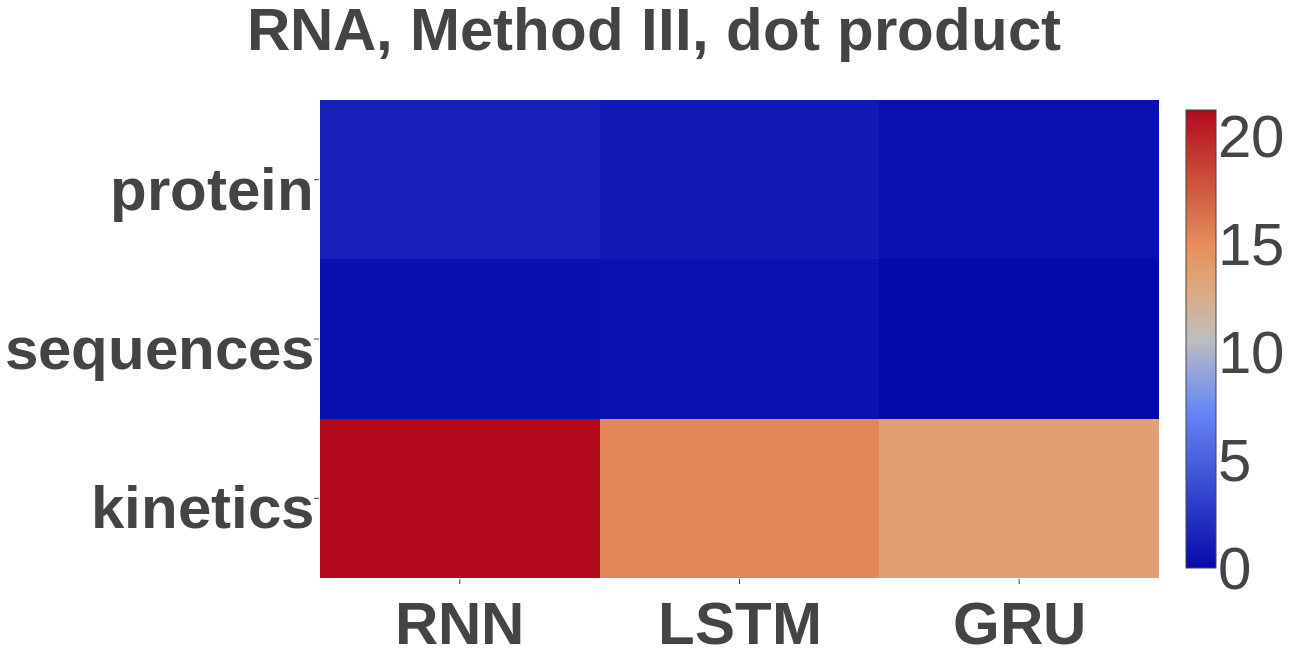}
  \caption{}
  \label{fig:do}
\end{subfigure}

\caption{Heatmaps showing the word scores fixing M\_LRC method and entities on JNLPBA dataset.}
\label{fig:dot_entity}
\end{figure*}


\textbf{Fixing an entity and a method}: Now, we fix a particular entity to analyze which model gives higher importance to different contextual words for a particular entity. Figure \ref{fig:dot_entity} shows the heatmaps by fixing entities $protein$, $DNA$ and $RNA$ respectively with M\_LRC method. ``protein'', ``sequences'' and ``kinetics'' have high frequency scores for $protein$, $DNA$ and $RNA$ respectively. The models capture this beautifully in all the cases.




\begin{table}[ht]
\centering
\begin{subtable}{.2\textwidth}
\scalebox{0.6}{
\begin{tabular}{|c|c|}
\hline
\textbf{Word}  & \textbf{Score} \\ \hline
(     &   9.407 \\ \hline
,     &   8.428 \\ \hline
ruling    &    2.537 \\ \hline
vice-chairman    &    1.41 \\ \hline
of    &    1.203 \\ \hline
national    &    0.901 \\ \hline
discuss     &   0.732 \\ \hline
congress    &    0.728 \\ \hline
the      &  0.723 \\ \hline
's      &  0.486 \\ \hline
minister  &      0.403 \\ \hline
and      &  0.209  \\ \hline
saturday    &    0.065  \\ \hline
0       & 0.03 \\ \hline
on     &   0 \\ \hline
friday   &     0 \\ \hline
)       & -0.002  \\ \hline
said     &   -0.023 \\ \hline
will    &    -0.045 \\ \hline
party    &    -0.068  \\ \hline
making    &    -0.072  \\ \hline
transparent&        -0.088 \\ \hline
efficient   &     -0.09  \\ \hline
foreign      &  -0.184 \\ \hline
more      &  -0.202  \\ \hline
\end{tabular}
}
\caption{}
\end{subtable}
\begin{subtable}{.2\textwidth}
\scalebox{0.6}{
\begin{tabular}{|c|c|c|}
\hline
\textbf{Word}  & \textbf{Score(Pr)} & \textbf{Score (CT)} \\ \hline
  control    &    0     &   0 \\ \hline
   and   &     -0.193   &     0  \\ \hline
  major    &    -0.487   &     -0.101  \\ \hline
  number    &    10.148   &     2.698 \\ \hline
 in    &    0.515   &     80.745 \\ \hline
  depressive    &    7.463    &    0.039 \\ \hline
  from    &    10.221    &    0.032 \\ \hline
  had    &    2.051     &   0.007 \\ \hline
  sites    &    -0.025     &   18.487 \\ \hline
 0    &    0     &   0 \\ \hline
   subjects   &     0   &     0 \\ \hline
  plasma   &     -0.083  &      0.001  \\ \hline
  recovered    &    -0.388     &   -0.014 \\ \hline
 cortisol    &    0.134   &     0 \\ \hline
  who   &     0.933   &     -0.002  \\ \hline
 measured    &    0.639    &    0.001  \\ \hline
  healthy     &   -0.047    &    0  \\ \hline
  of     &   36.08    &    4.335  \\ \hline
  dgdg    &    -0.343    &    -0.001  \\ \hline
 patients   &     3.377  &      0.007  \\ \hline
  were      &  0.454   &     0.001  \\ \hline
  concentrations &       0.014    &    0  \\ \hline
  the    &    -0.613   &     2.572 \\ \hline
 disorder  &      10.723 &       0  \\ \hline
\end{tabular}
}
\caption{}
\end{subtable}

\caption{Entity wise relevance scores for words in two individual sentences using LSTM model: \textbf{(a)} Using M\_SLL method for CoNLL instance and \textbf{(b)} Using M\_LRC method with dot product for JNLPBA instance.}
\label{fig:lstm_dot_sent}
\end{table}
At a sentence level, we only consider our best performing model, LSTM. Table \ref{fig:lstm_dot_sent} gives entity wise word relevance scores for two individual sentences. It uses a sentence from CoNLL dataset - ``Saturday 's national congress of the ruling Czech ({\it I-ORG}) Civic ({\it I-ORG}) Democratic ({\it I-ORG}) Party ({\it I-ORG})  ODS ({\it I-ORG})) will discuss making the party more efficient and transparent , Foreign Minister and ODS ({\it I-ORG}) vice-chairman Josef ({\it I-PER}) Zieleniec ({\it I-PER}), said on Friday .''. The tags for all entity words are mentioned alongside each word. Notice the high scores for ``vice-chairman'', ``ruling'', ``congress'', ``minister'' meets the intuitive understanding of these words. Interestingly, round brackets get the maximum scores for M\_SLL method, which may be attributed to their frequent use with $ORG$ entity words. Similarly, sentence taken from JNLPBA dataset is: ``the number of glucocorticoid ({\it B-protein}) receptor ({\it I-protein}) sites in lymphocytes ({\it B-cell_type}) and plasma cortisol concentrations were measured in dgdg patients who had recovered from major depressive disorder and dgdg healthy control subjects .''. Again, higher scores for ``sites'' and ``plasma'' for $cell\_type$ are in agreement with overall scores given to them.
\subsection{Positional effects of context words}
\begin{table*}[t]
\begin{minipage}{\textwidth}
\centering
\scalebox{0.9}{
\begin{tabular} 
{|p{0.08\linewidth}|p{0.08\linewidth}|p{0.08\linewidth}|p{0.7\linewidth}|} \hline
{\textbf{RNN}} & {\textbf{LSTM}} & {\textbf{GRU}} & \multirow{1}{*}{\textbf{Sentence}} \\ \cline{1-4}
0.0 & 0.0 & 0.0 & Senegal proposes foreign {\bf minister} for U.N. post . \\ \hline
0.163 & 2.576 & 1.031 & He was senior private secretary to the employment and industrial relations {\bf minister} from 1983 to 1984 and was Economic advisor to the treasurer Paul Keating in 1983 .\\ \hline
239.793 & 112.405 & 199.985 & The ODS , a party in which Klaus often tries to emulate the style of former British Prime {\bf Minister} Margaret Thatcher , has been in control of Czech politics since winning general elections in 1992 \\ \hline
\end{tabular}
}
\caption{Relevance scores for the word ``minister''  in three different test sentences from CoNLL dataset.}
\label{tab:minister_pos}
\end{minipage}
\end{table*}

\begin{figure}[h]
\begin{subfigure}{.5\textwidth}
  \centering
  \includegraphics[width=\linewidth]{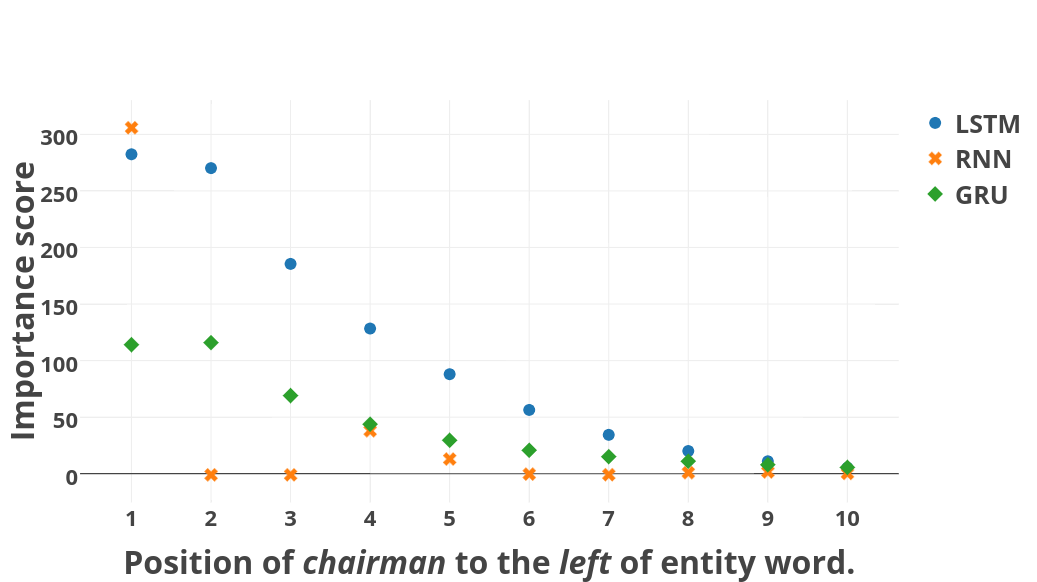}
  \caption{}
  \label{fig:chairman}
\end{subfigure}
\begin{subfigure}{.5\textwidth}
  \centering
  \includegraphics[width=\linewidth]{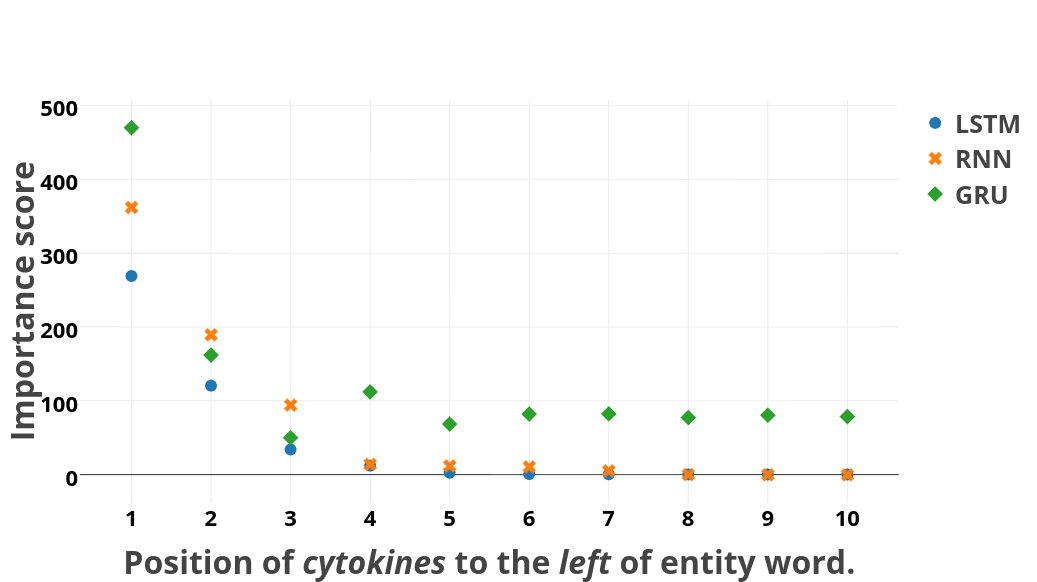}
  \caption{}
  \label{fig:cytokines}
\end{subfigure}

\caption{Position vs relevance score plot for three models for (a) ``chairman'' w.r.t. $PER$ entity word ``Josef'' and (b) ``cytokines'' w.r.t. $protein$ entity word ``erythropoietin''.}
\label{fig:position_plots}
\end{figure}

In this section, we analyze how the position of context words affects their scores obtained by M\_LRC method. We do this analysis for real sentences present in the test sets as well as on artificial sentences. We achieve this by applying the proposed techniques at an individual sentence level. For instance, Table \ref{tab:minister_pos} shows the relevant scores of the word ``minister'' for entity $PER$ obtained by three models, in three test sentences taken from CoNLL dataset. M\_WF method indicates that ``minister'' has high importance for entity type $PER$ matching with our intuition. However ``minister" is likely to appear in different sentences with different context and may not have equal relevance as also indicated in the Table~\ref{tab:minister_pos}. 
In the first sentence, there is no entity word for $PER$, hence, the score for ``minister'', corresponding to entity $PER$ is zero. In the second sentence, the score is higher, though not too high as the word is relatively far from the relevant entity word. However, the score is much higher in the third sentence where ``minister'' is right before the entity words ``Margaret Thatcher''. Relative scores obtained by using different neural models also match with the general notion that RNN tends to forget long range context (second sentence) compared to LSTM and GRU, and is quite good for short distance context (third sentence).

We further validate the above observation on artificial examples. Figure \ref{fig:chairman} gives the position verses score plot for the word ``chairman'' with respect to the $PER$ entity word ``Josef''. The position tells that how far to the left ``chairman'' is from the entity word. We create sentences as follows - ``chairman Josef .'', ``chairman R Josef .'', ``chairman R R Josef .'' and so on. Here, R represents a random word. It can be observed that how LSTM and GRU assign a higher score to far off words compared to RNN, justifying their ability to include such words in making the final decision. 

Figure \ref{fig:cytokines} shows a similar plot for the word ``cytokines'' and a $protein$ entity word ``erythropoietin'' using the same way of creating artificial sentences. Interestingly, GRU assigns higher relevance scores than LSTM and RNN, which is in accordance with the high overall score it gives to ``cytokines'' compared to the other two models.

\begin{table}[h]
\centering
\scalebox{0.7}{
 \begin{tabular}{|c | c | c|} 
 \hline
 {\textbf{Rank}} & {\textbf{Word}} & {\textbf{Score}} \\ [0.5ex] 
 \hline\hline
1 & by & 66.162 \\
2 & the & 22.223 \\
3 & in & 3.576 \\
4 & expression & 0.257 \\
5 & can & 0.222 \\
6 & gene & 0.221 \\
7 & which & 0.079 \\
8 & over & 0.079 \\
9 & important & 0.003 \\
10 & may & 0.002 \\
11 & establishing & 0 \\
12 & type & 0 \\
13 & cell & 0 \\
14 & 0 & 0 \\
15 & specificity & 0 \\
16 & and & 0 \\
17 & widening & -0.001 \\
18 & range & -0.016 \\
19 & recognized & -0.364 \\
20 & be & -0.475 \\
21 & modulated & -0.534 \\
22 & degeneracy & -0.857 \\
23 & sequences & -0.917 \\ [1ex]
 \hline
 \end{tabular}
 }
\caption{Relevance scores for an individual test sentence from JNLPBA dataset, using LSTM and M\_LRC method with dot product.}
\label{table:file2}
\end{table}

\subsection{Error Analysis}
The proposed methods can be effectively used to conduct error analysis on bi-directional recurrent neural network models. For a given sentence, a negative score for a particular word means that the model is able to make a better decision when the word is removed from the sentence. Relevance scores can be used to find out which words confuse the model. Knowing what those words are, is crucial to understanding why the model makes a mistake in a particular instance. For example, Table \ref{table:file2} shows the word importances for the sentence - ``the degeneracy in sequences recognized by the otfs ({\it B-Protein}) may be important in widening the range over which gene expression can be modulated and in establishing cell type specificity .'' The LSTM model makes a mistake here by tagging ``otfs'' with tag {\it B-DNA}. Words ``degeneracy'', ``sequences'', ``widening'', ``recognized'' and ``modulated'' all have a higher overall score for $DNA$ entity class than for $protein$. Hence, the presence of these words in the sentence fool the model into making a wrong decision.

In general, we observe that the presence of words which have high scores for false entity types tend to confuse the model. Position of words also plays a vital role. Words which appear in a far off or a different position than what they generally appear in the training dataset, tend to receive negative or low scores even if they are important. For instance, ``minister'' mostly appears to the left of an entity word in the training dataset. If, in a test case, it appears to the right, it ends up receiving a low score.
   
\section{Related Work}
Various attempts have been made to understand neural models in the context of natural language processing. Research in this direction can be traced back to~\newcite{elman1989representation} which gains insight into connectionist models. This work uses principal component analysis (PCA) to visualize the hidden unit vectors in lower dimensions. Recurrent neural networks have been addressed in recent works such as~\newcite{karpathy2015visualizing}. Instead of a sequence tagging task, they use character level language models as a testbed to study long range dependencies in LSTM networks.

~\newcite{li2015visualizing} build methods to visualize recurrent neural networks in two settings: sentiment prediction in sentences using models trained on Stanford Sentiment Treebank and sequence-to-sequence models by training an autoencoder on a subset of WMT'14 corpus. In order to quantify a word's salience, they approximate the output score as a linear combination of input features and then make use of first order derivatives. Erasure technique helps us to do away with such assumptions and find word importances in sequence labeling tasks for individual entities.

Similar to present work, ~\newcite{kadar2016representation} analyze word saliency by defining an omission score from the deviations in sentence representations caused by removing words from the sentence. This work, however, targets a different, multi-task GRU framework, learning visual representations of images and a language model simultaneously. 

Another closely related work is~\newcite{li2016understanding}. They use erasure technique to understand the saliency of input dimensions in several sequence labeling and word ontological classification tasks. Same technique is used to find out salient words in sentiment prediction setting. Our work focusing on sequence labeling task has several differences with~\newcite{li2016understanding}. Firstly, in case of sequence labeling,~\newcite{li2016understanding} only focus on feed forward neural networks while our work trains three different recurrent neural networks on general and domain specific datasets. Secondly, their analysis in sequence labeling task is only limited to important input dimensions. Instead, our work focuses on finding salient words which are basic units for most NLP tasks. Lastly, our M\_SLL method is an adaptation of their method to find salient words in sentiment prediction task. Unfortunately, for a sequence labeling task, this method is not very suitable. Since it only considers sentence level log likelihood, it makes no distinction between various possible entities such as person or organization. Our M\_LRC method, which takes individual word level effects into account, is more suitable.

A significant amount of work has been done in Computer Vision to interpret and visualize neural network models~\cite{simonyan2013deep,mahendran2015understanding,nguyen2015deep,szegedy2013intriguing,girshick2014rich,zeiler2014visualizing,erhan2009visualizing}. Attention can also be useful in explaining neural models~\cite{bahdanau2014neural,luong2015effective,sukhbaatar2015end,rush2015neural,xu2016ask}.
\section{Conclusions and Future Work}
In this paper, we propose techniques using word erasure to investigate bi-directional recurrent neural networks for their ability to capture relevant context words. We do a comprehensive analysis of these methods across various bi-directional models on sequence tagging task in generic and biomedical domain. We show how the proposed techniques can be used to understand various aspects of neural networks at a word and sentence level. These methods also allow us to study positional effects of context words and visualize how models like LSTM and GRU are able to incorporate far off words into decision making. They also act as a tool for error analysis in general by detecting words which confuse the model. This work paves the way for further analysis into bi-directional recurrent neural networks, in turn helping to come up with better models in the future. We plan to take our analysis further by including other aspects like character and word level embedding into account.

\bibliography{acl2016}
\bibliographystyle{acl2016}


\end{document}